\documentclass{article}

\usepackage[preprint]{nips_2018}
\usepackage[utf8]{inputenc} 
\usepackage[T1]{fontenc}    
\usepackage{url}            
\usepackage{booktabs}       
\usepackage{amsfonts}       
\usepackage{nicefrac}       
\usepackage{microtype}      
\usepackage{graphicx}
\usepackage{amsmath}
\usepackage{hyperref}       

\title{{GeoGAN: A Conditional GAN with Reconstruction and Style Loss to Generate Standard Layer of Maps from Satellite Images}}

\author{
    Swetava Ganguli\thanks{Corresponding Author}, Pedro Garzon, Noa Glaser \\
    Department of Computer Science\\
    Stanford University, Stanford, CA 94305 \\
    \texttt{\{swetava,pgarzon,nglasr\}@cs.stanford.edu}
}
  
\begin{document}

\maketitle

\begin{abstract}
    Automatically generating maps from satellite images is an important task. There is a body of literature which tries to address this challenge. We created a more expansive survey of the task by experimenting with different models and adding new loss functions to improve results. We created a database of pairs of satellite images and the corresponding map of the area. Our model translates the satellite image to the corresponding standard layer map image using three main model architectures: (i) a conditional Generative Adversarial Network (GAN) which compresses the images down to a learned embedding, (ii) a generator which is trained as a normalizing flow (RealNVP) model, and (iii) a conditional GAN where the generator translates via a series of convolutions to the standard layer of a map and the discriminator input is the concatenation of the real/generated map and the satellite image. Model (iii) was by far the most promising of three models. To improve the results we also added a reconstruction loss and style transfer loss in addition to the GAN losses. The third model architecture produced the best quality of sampled images. In contrast to the other generative model where evaluation of the model is a challenging problem. since we have access to the real map for a given satellite image, we are able to assign a quantitative metric to the quality of the generated images in addition to inspecting them visually. While we are continuing to work on increasing the accuracy of the model, one challenge has been the coarse resolution of the data which upper-bounds the quality of the results of our model. Nevertheless, as will be seen in the results, the generated map is more accurate in the features it produces since the generator architecture demands a pixel-wise image translation/pixel-wise coloring. A video presentation summarizing this paper is available at: \url{https://youtu.be/Ur0flOX-Ji0}
\end{abstract}

\section{Introduction}
 Creating maps is a very expensive and time consuming process; yet it is one of the most important sources of curated data. Maps have commercial value to companies in multiple sectors of the economy: ride-sharing companies (like Uber and Lyft), food delivery companies like (like DoorDash and GrubHub), national security agencies (like CIA, NSA, and FBI), and many other sectors of the economy. It is important to emphasize that in addition to the map being accurate, most of the apps created by the companies mentioned above have a major human-interaction component associated with their success. For e.g. Customers interact with a map interface while tracking their ride on the Uber/Lyft app while apps that geo-code photographs taken on a device present the pictures associated with a given geographic location on a human-readable map interface. Constructing accurate maps has been a major push for companies that want to sell ``smart devices'' since a major part of what makes a device smart is its ability to locate itself and inform its user via a human-readable interface. More importantly, maps have a huge humanitarian value. For e.g., generating an accurate map of refugee camps can aid the timely delivery of food and resources to the people in these camps. A map may be also viewed as data that is not necessarily machine readable. This approach is suitable in the domain of autonomous vehicles and self-driving cars. An accurate map must reflect all changes on the ground in a timely manner. Up-to-date geospatial data is continuously collected from flyover imaging air-crafts or satellites. Currently, there is considerable latency between changes to geographic/road conditions on the ground and the publicly available human-readable maps. One way to reduce this latency is to automate the process of human-readable map generation from a satellite image of a given location at a specified zoom level and resolution. In this work, we emphasize the importance of human-readability of a map and aim to construct accurate human-readable maps directly from a satellite/aerial image of the location. The satellite/aerial image specifies the zoom level and resolution of the required map. The style of the generated human-readable map is chosen to be the style of the publicly available Google Maps API. Satellite images are obtained from the publicly available Google Earth Engine datasets and its corresponding map is obtained using the Google Maps API. The process of querying these databases and aligning the satellite/aerial image with its corresponding map is described in the section on datasets.

\section{Related work}
Image-to-Image translation has been a recent development and area of research in the field of generative modeling. One of the most recent successes is the work of \cite{chan2018dance}. The use of GANs (\cite{Goodfellow14}) for novel generative tasks has exploded over the past few years due to better training algorithms (\cite{Gulrajani17}) and novel ways to define the loss function which must be optimized (\cite{Arjovsky17}). \cite{Costea17} proposed the dual-hop GAN (DH-GAN) to detect roads and intersections and used that to find its best covering road graph by applying a smoothing-based graph optimization procedure. In addition, using convolutional neural networks and GANs on geospatial data in unsupervised or semi-supervised settings has also been of interest recently; especially in domains such as food security, cybersecurity, satellite tasking, etc. (\cite{ganguli2019predicting, perez2019semi, dunnmon2019predicting}). Conditional Adversarial Networks (\cite{Mirza14,Isola16}) have been used to perform general purpose image-to-image translation. The recently proposed CycleGAN architecture has been evaluated on satellite imagery to map translation (\cite{zhu18}). GANs have been also used to create spoof satellite images and spoof images of the ground truth conditioned on the satellite image of the location (\cite{Xu18}). Conditional GANs have also been used to generate ground-level views of locations from overhead satellite images \cite{Deng18}. 

As an alternative to standard neural network generators, there has also been progress and recent models proposed based on normalizing flow models that use invertible bijections of distributions to convert one distribution to another and back again. An alternative architecture for the generator in a GAN would be a normalizing flow model which takes advantage of the bijective mapping between the input noise/condition vector and the generated image and could in principle be used to gain more insight by training the model to seek disentangled latent representations of the map. Recently proposed normalizing flow models that have succeeded in face generation is the Real-Valued Non-Volume Preserving Normalizing Flow Model (RealNVP). The RealNVP has been shown to have decent success in the task of photorealistic face generation
(\cite{DinhSB16}) and disentangling its latent representations. In this project, we modify the conditional GAN architecture in various ways to obtain the one best suited for our task of generating a map for a location given the satellite/aerial image of the location. In addition to experimenting with new architectures, we also analyze the importance of different flavors of loss functions that penalize the model on certain specified quantitative metrics of generated sample quality to get a more comprehensive survey of the task. We also contribute code that provides a scalable method to generate new datasets from the publicly available Google Earth Engine and the Google Maps APIs.

\section{Task Definition}
GeoGAN is a model that takes as its input a satellite image at a specified zoom level and resolution and produces the corresponding human-readable map for that location. The model is trained to emulate the style of the maps available from the Google Maps API for the generated map. The emphasis in this paper is on the accuracy of the generated map. As opposed to many generative tasks, we have access to the real map corresponding to a satellite image which is obtained from the Google Maps API. The model is trained as a conditional GAN with the input satellite image as the condition for the generation of the map. While it is relatively straightforward to condition the output of a generator, there is no clear way to condition the output of the discriminator. We perform several experiments on the methods to condition the discriminator. For most of our experiments, our inputs are 64x64x3 (3 channel) satellite images which are translated into 64x64x3 map images. However, for the model christened ``model 3'', we use 256x256x3 dimensional satellite images to generate 256x256x3 maps.

\subsection{Datasets}
We need tiles with corresponding spatial resolution at locations at specific (latitude, longitude) pairs from street maps, overhead imagery, and geospatial elevation datasets. We built pipelines to download the images from Google Earth engine and the Google Maps API. Our pipeline first calculates a set of lat/long coordinates to span a given city. These lists of coordinates are then fed into scripts downloading data from Google Earth Engine and Google Maps. Then, a third script matches the images and stores them as Tensorflow data records containing corresponding satellite images and map tiles. So far we have extracted images from San Francisco, Seattle, New York, Austin, Miami, Washington D.C., Boston, and Chicago for a total of 41756  for both training and test, of which we cleaned out 7605. For satellite imagery, we will be using Sentinel2 from Google Earth Engine, which has a 10m pixel resolution over the red/green/blue NIR bands \cite{Sentinel2}. A benefit of using this data source for satellite images is that querying is free and you also get access to other channels such as an infrared red channel. It also contains other views that better topographic representation of a location. Because the ground looks different in different parts of the year, we download images from around March, June, September, and December from each region. We used the provided cloud-mask channel to select the images with the least clouds. If no images in a month contain less than 10\% cloud coverage, we extend the dates region. For satellite imagery, we also have access to higher resolution digital globe data from the DeepGlobe challenges. However we are not using this data at the moment. The roads challenge dataset, for example, consists of 6226 satellite images of 1024 x 1024 pixels with a 50cm pixel resolution in the RGB bands \cite{DeepGlobe}. This fine resolution is great and future work might explore incorporating this data. For the maps data, we will be using tiles from the Google Maps Static API. This API allows us to retrieve tiles from anywhere in the globe without street labels and at a few specific discrete zoom levels. We have opted to use zoom level 14, which roughly translates to 7.242090 meters per pixel resolution as 512 x 512 PNG images. We form our tiles by going to the top left corner of a city all the way to the bottom right corner. In addition to our handcrafted dataset, we also compare using the collected dataset from UC Berkeley researcher Taesung Park that uses the Google Maps API to scrape both satellite and map images of size 256 x 256 from New York City \cite{park_dataset}.

\begin{figure}
    \vspace{1cm}
	\centering
	\includegraphics[width=0.7\textwidth]{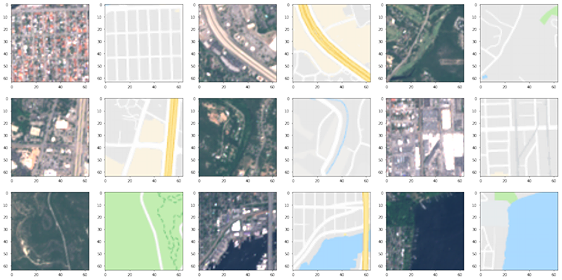}
    \caption{A grid comparing the the satellite images from Google Earth Engine along with the corresponding Google Maps tile}
    \label{training_data}
\end{figure}

\section{Technical Approach}
\subsection{Our Contributions}
\cite{zhu18} present the CycleGAN architecture which  translates satellite imagery to maps. In this paper, we experiment with and propose alternate novel architectures and loss functions to train for this task. 

\begin{enumerate}
    \item A Conditional GAN emulating the architecture proposed by \cite{Mirza14} where the satellite images are first passed through an encoder to compress them to a lower dimensional vector. We experiment with both, an approach where the encoder is trained end-to-end, and an approach where the encoder is pre-trained over satellite images as part of an auto-encoder. We also experiment with using varying amounts of noise as either a concatenation to the encoded vector or an addition to the encoded vector.
    \item Adding a pixel-wise reconstruction loss in addition to the GAN loss which measures the pixel-wise normed distance between the generated maps and the real map which is available at training time.
    \item Adding a style reconstruction loss to the  losses described in the previous item. The style reconstruction loss has the same mathematical structure as the style loss in neural style transfer \cite{Gatys15}. This loss emphasizes not only pixel-wise coherence but also makes sure that the corrlation between a pixels and their neighbors have the same characteristics as the real map.
    \item 
    Experimenting with performance using L1 versus L2 norm for determining the reconstruction loss. The different norm operations have been known to aid training depending on the task. While CycleGAN (\cite{zhu18}) benefits from the L1 penalty in its cycle-consistency loss, we observe that the L2 norm performs better for our task.
    \item While it is straightforward to condition the output of the generator of a GAN to a given input, it is not clear how to condition the discriminator. We try two approaches. The first approach is to use the compressed and encoded version of the satellite image and concatenate it to the prefinal layer of the output of the discriminator before passing it through a final set of layers to obtain the discriminator's decision probability. The other approach is to concatenate the satellite image to the real or the fake image along the channel axis. This allows the convolution filters to have access to the full satellite image while making a decision on how realistic the generated maps are. This approach therefore totally bi-passes the need for an encoder. 
    \item Using a RealNVP generator instead of the standard convolutional generator. RealNVP is a bijective model and thus allows us to translate very efficiently from the map image domain to the satellite image domain and the vice versa.
    \item While we are continuing to work on increasing the accuracy of the model, one challenge has been the coarse resolution of the data which upper-bounds the quality of the results of our model. Nevertheless, as will be seen in the results, the generated map is more accurate in the features it produces since the generator architecture demands a pixel-wise image translation/pixel-wise coloring. For example, if we look at tile number 6 in figure 6, we see that the generated map is more accurate and has more features than the real map when compared with the available satellite image.
    \item To the best of our knowledge, ours is the first model to incorporate a reconstruction loss (for pixel-wise accuracy) and a style loss (to reduce high frequency artifacts) in addition to the GAN loss (a feature-wise learnt similarity metric or content loss similar to the ideas presented in \cite{Gatys15} and \cite{LarsenSW15}) for the task of generating the standard layer of the map from a satellite image.
\end{enumerate}

\subsection{Evaluation Method}
Given that GANs are not evaluated by tracking likelihoods, most of our evaluation will be qualitative. I.E.  to see if our models learned the distribution of well, we can sample new satellite images, generate maps for them, and perform a small experiment to see whether humans and can differentiate between the real and generated maps/topographies. However, we do have "ground truths" for the map translation task and can also track image similarity indexes. Some similarity indexes include L1 and L2 reconstruction losses. We track a combination of L2 loss and style reconstruction loss and see an improvement over training. We created three high level models based on three different philosophies which are described below.

\subsection{Model 1: Conditional GAN}

\begin{figure}
	\centering
	\includegraphics[width=0.8\textwidth]{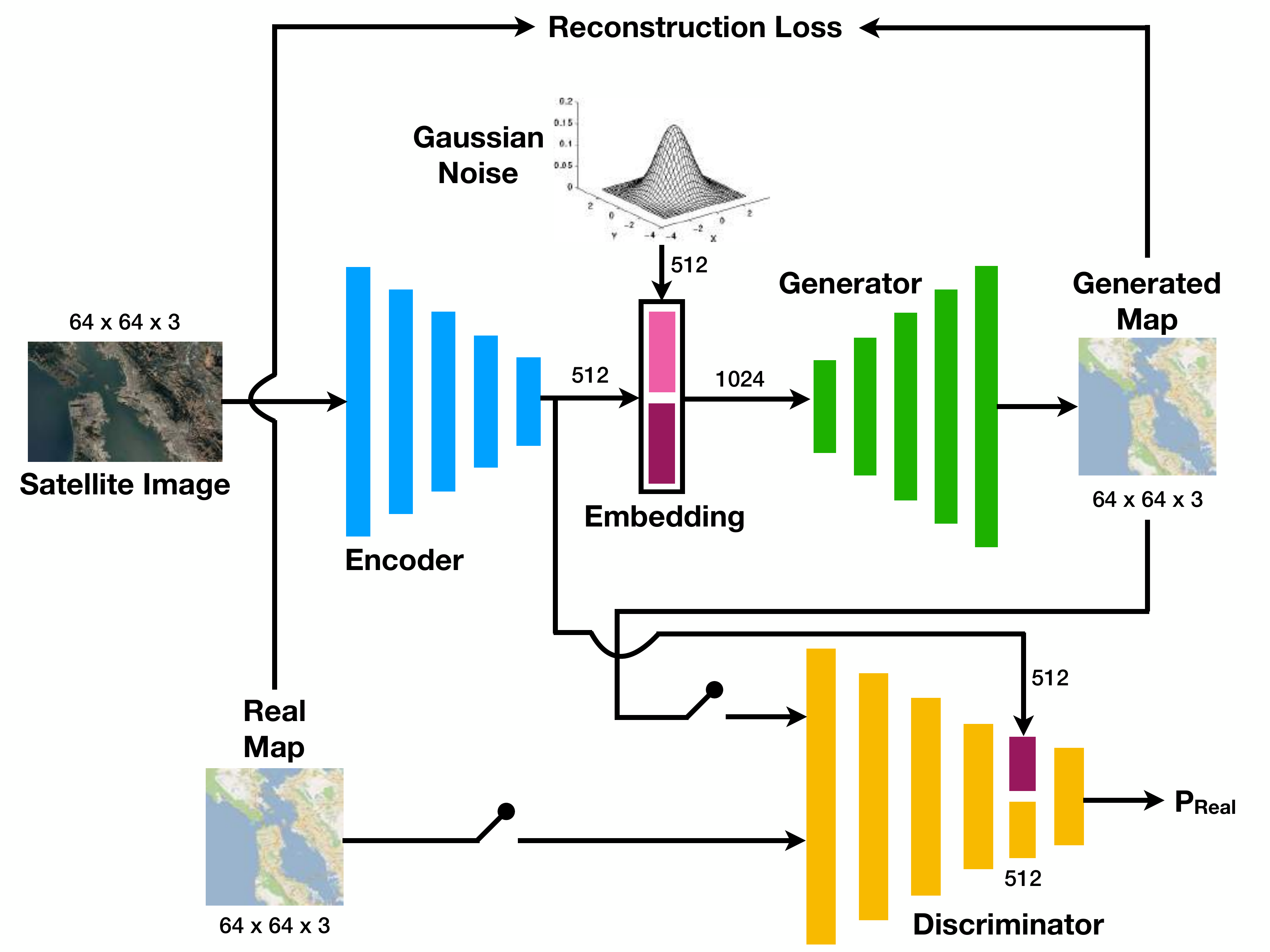}
    \caption{A schematic of the baseline GeoGAN model. The generation of the map is conditioned on the satellite image along with some optional noise. The discriminator is also supplied with the conditional information from the satellite image embedding. One architecture we used is as follows (with c($x$ x $y$ x $z$) s$n$ referring to a convolution with a $z$ filters of size $x*y$ and a stride of $n$, cT($x$ x $y$ x $z$) s$n$ is the same with transposed convolutions, and lr referring to a a leaky rely layer): \textbf{Encoder}: input=batchx64x64x3; layers = c(3x3x1024) s2, lr, c(3x3x512) s2, lr  c(3x3x256) s2, lr, c(3x3x128) s2, lr, c(4x4x512) s1, batchnorm, lr; output= c(batchx1x1x512) \textbf{Noise}: input=batchx64x64x3; element wise add noise; output=batchx64x64x3 \textbf{Generator}: input=batchx1x1x512; layers cT(4x4x1024) s1, lr, cT(8x8x512) s2, lr, cT(3x3x128) s2, lr, cT(3x3x32) s2, lr, cT(3x3x64) s2; output=batchx64x64x3 \textbf{Discriminator}: input=batchx64x64x3; c(3x3x128) s2, lr, c(3x3x256) s2, lr, c(3x3x512) s2, lr,  c(3x3x1024) s2, lr, c(4x4x512) s1, lr, concatenate 512 activation units with embedding of sat image, fully connected layer with 512 outputs, lr, fully connected layer, sigmoid. Convolution padding is omitted for brevity. \label{DCGAN_Model}}
\end{figure}

One way to think about our task is that want the Generator, instead of generating a map image based random noise, to generate a map image based on information in the satellite tile. Creating a lower dimensional representation of the satellite images to feed into the generator will allow could afford several very nice tasks such as (1) Analyze how the generator treats this lower dimensional space, hopefully finding disentangled representations (2) More naturally interpolate between different satellite images in this space (3) More naturally add custom features and hope that the generator uses them in order to later interpolate them and create new maps with tweaked high level attributes (ie. population density in the region, the lat/long region, the city, the year, etc.). \cite{Mirza14} was able to generate encouraging results for GAN Generators trained on feature vectors, and so model 1 is constructed to extract the useful information from the satellite image into a feature vector, then pass it to a generator. 64x64x3 are passed through a deep convolutional network (Encoder) to create 100 or 512 long embeddings. These embeddings are either appended or element wise added Gaussian noise. Then a convolutional Generator is provided these embedding to generate the 64x64x3 map. The Encoder and Generator are trained jointly with a convolutional Discriminator using the standard GAN discriminator loss. The standard GAN non-saturating generator loss is used for the generator loss. Given we know the correct fake-real map pairs, a reconstruction loss can be computed as the norm of the difference between the images. Figure \ref{DCGAN_Model} shows the model architecture. 

\textbf{Discriminator loss}
\begin{equation} \label{GAN_Loss}
\max_D \mathbb{E}_{\textbf{x} \sim \textbf{p}_{\textrm{Real}}}
\left[\log D\left(\textbf{x}\right)\right] + 
\mathbb{E}_{\textbf{z} \sim \textbf{p}_\textbf{Noise}}[\log (1-D(G(\textbf{z})))] 
\end{equation} 

\textbf{Generator loss}
\begin{equation}\label{GAN_nonsaturating_Loss}
	\min_G - \mathbb{E}_{\textbf{z} \sim \textbf{p}_\textbf{Noise}}  [\log D(G(\textbf{z}))] 
\end{equation}

\textbf{Reconstruction loss}
\begin{equation}\label{reconstruction_loss}
|| M_{\text{fake}} - M_{\text{real}} ||_2
\end{equation}

Embedding the input in a lower dimensional vector and appending noise was inspired by \cite{Deng18}.
Using deep fully convolutional networks was motivated by \cite{Radford15}. We experimented with different embeddings such as 100 and 512 long vectors, different ways to add noise (appending vs adding element wise), different relative lengths and magnitudes of the noise vectors, and both L1 and L2 reconstruction losses.

\subsection{Model 2: Conditional GAN without Encoder and with reconstruction loss}
\begin{figure}
	\centering
	\includegraphics[width=0.8\textwidth]{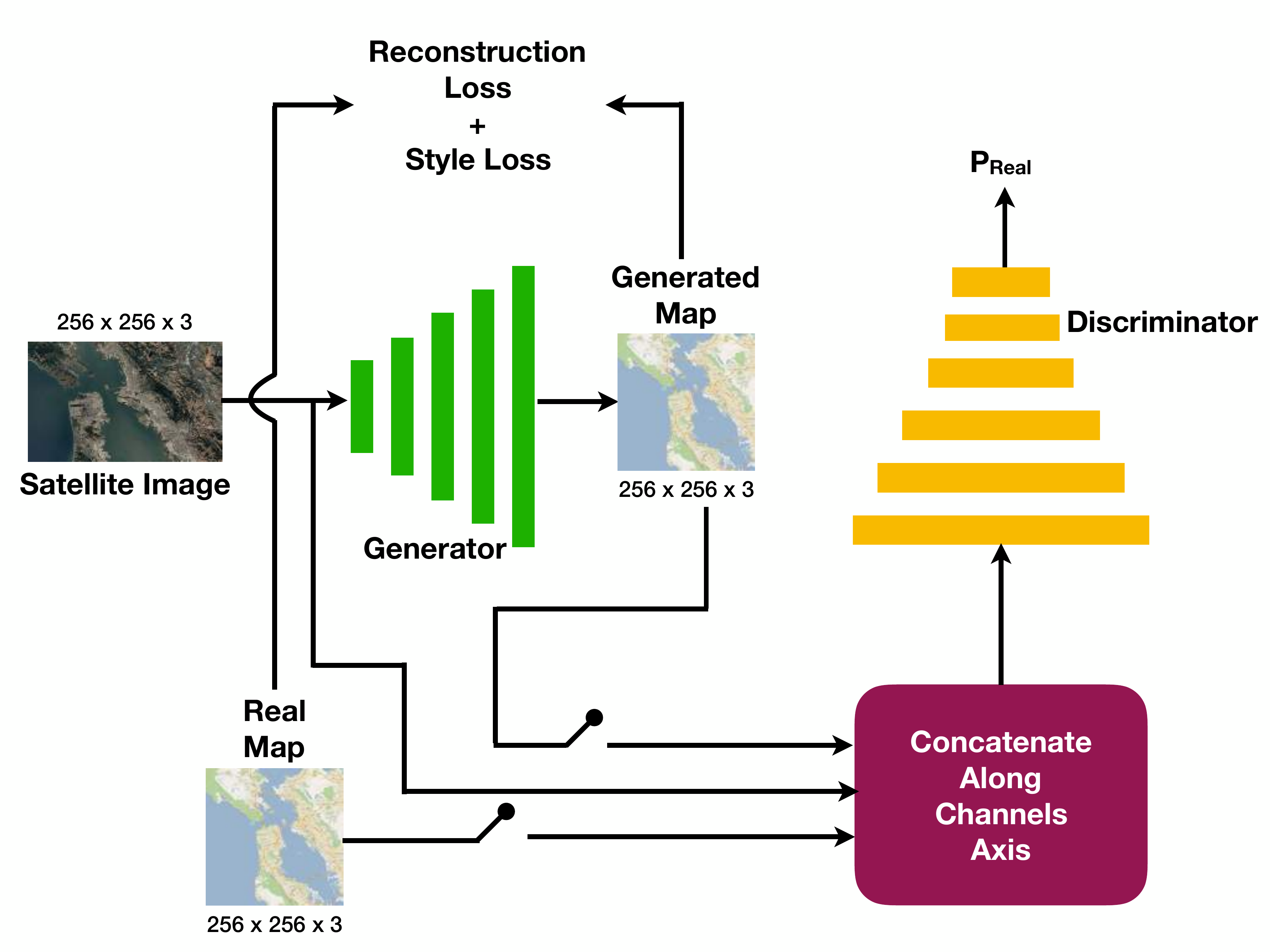}
    \caption{A schematic of the GeoGAN model without encoding. The generation of the map is conditioned on the entire satellite image. Each input to the discriminator is 256 x 256 x 6, which is a satellite image concatenated with either the real or generated map along the channel axis. A more detailed description of our architecture is below: \textbf{Generator:} To give two different sizes of receptive fields a chance to compete for weights in each layer, each convectional layer is a channel wise concatenation of n (3x3) convolutional filters and n (5x5) convolutional filters followed by a leaky ReLu. For the first four layers, n is equal to 300,150,60, and 20. The last layer is a transposed 3x3 convolution bringing the last 40 layers to 3 followed by a tanh.\textbf{Discriminator:} The first four layers in the discriminator are 3x3 convolutions with stride two and 128, 256, 512, 768 and 1024 filters respectively. (All followed by leaky ReLu layers). The last layer is a 4x4 convolution followed by sigmoid. \label{DCGAN_no_encoder_Model}}
\end{figure}

The second approach for satellite to image translation involves treating it as a cross between an image generation and a pixel level discrimination between the different map color classes (gray for road, green for vegetation, blue for water, etc.). We could use a U-net for this model, as is done by  \cite{Isola16}, which uses a U-net with a PathGAN discriminator (the PatchGAN penalizes textures of the output that are too different from those in ground truths. However, the simplest model is just a series of convolutions which give every pixel an effective receptive field around it's position in the original image. As such we came up with the architecture shown in Figure \ref{DCGAN_no_encoder_Model} which does not include an encoder. We also added a reconstruction loss (equation \ref{reconstruction_loss}). We noticed that some of the details from the satellite images are transferred over to the generated maps, so we added a style transfer loss. The style-transfer loss can be written as follows where $G_{ij}$ represents the gram-matrix for the generated image and $A_{ij}$ is the gram-matrix for the ground truth maps image. The gram-matrix represent the inner-product of the image $F$. $N$ represents the amount of channels and $M$ is the product of the height and width of the images.

\textbf{Style loss}
\begin{equation}\label{style_loss}
\begin{aligned}
& G_{ij} = \sum_{k} F_{ik} F_{jk}\\
& L_{style} = \frac{1}{4 N^2 M^2} (G_{ij} - A_{ij})^2
\end{aligned}
\end{equation}
In the original style transfer paper, this style loss is calculated as a sum over the activation of all the layers. However, in our case we don't have a completely trained model to extract style from, so we suffice with taking the style loss of the map images directly.

\subsection{Model 3: Real NVP}
\cite{zhu18} showed impressive results \cite{zhu18} on satellite to map translation using convolutional Cycle GANS. In their paper, the networks for the forwards and backwards (satellite \(\rightarrow\) map, and map\(\rightarrow\)satellite) were trained jointly using a reconstruction loss. Flow models have the benefit of being invertible so they make natural fits for cycle GANs. In more detail, a flow model and it's inverse implicitly incorporates a direct reconstruction without having to make the model learn such through adding additional loss penalties or a cycle-consistency loss. The forward direction is simply the bijection from a noise distribution to the image, and the backward direction will be its inverse. Because the forward and backward operations are exact inverses of each other, there should be no reconstruction loss, therefore the loss function should be the sum of the two discriminator's losses. We chose RealNVP for our flow model because it is has shown good performance on more complicated real world images such a faces and is not as hard to train as Glow \cite{Kingma18}. We attempt creating the RealNVP architecture similarly to \cite{DinhSB16} by first using 2 bijections with checkerboard masking followed by another 2 bijections channel-wise masking. The different masking operations are to vary which half of the total pixels get transformed by the RealNVP bijection. All bijections use the RealNVP bijection implementation in Tensorflow with two FC layers of 512 units along with batchnorm. This method differs as a more simple implementation than the full RealNVP that uses a residual based network. It should also be noted that the RealNVP model used in the original paper trains through maximum likelihood optimization, which is what a GAN based approach tends to avoid  (\cite{DinhSB16}).

\section{Results}

\subsection{Conditional GAN with encoder}
\begin{figure}
	\centering
	\includegraphics[width=\textwidth]{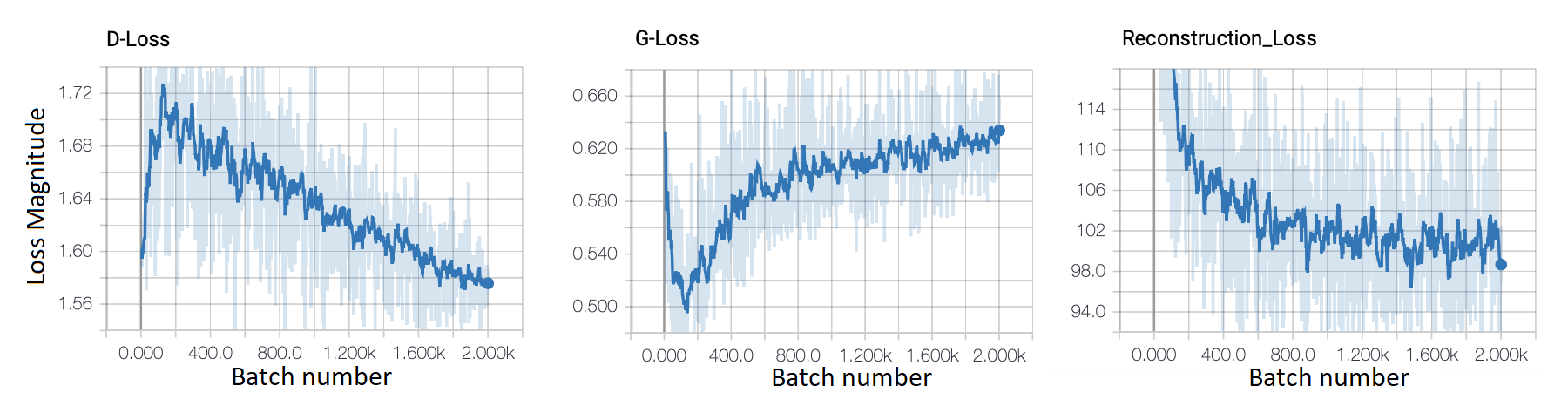}
    \caption{The generator, discriminator and reconstruction losses of the conditional GAN with the encoder and only the reconstruction loss trained on the dataset we generated. The discriminator is very quickly learning to tell real maps from fake maps and the Generator and discriminator losses saturate at their current points. \label{encoded_GANS_Losses}}
\end{figure}

While this model performed well on our MNIST sanity check (see figure \ref{MNIST_Generated} in the appendix) it failed to train well on our satellite images dataset. The training losses would saturate for both the Generator and Discriminator (see figure \ref{encoded_GANS_Losses}). The samples were also bad quality. This seemed to always be the case when we tried changing the noise levels noise at 0\%, 10\%, and 50\% noise.

\subsection{Conditional GAN without encoder}
\begin{figure}
	\centering
	\includegraphics[width=\textwidth]{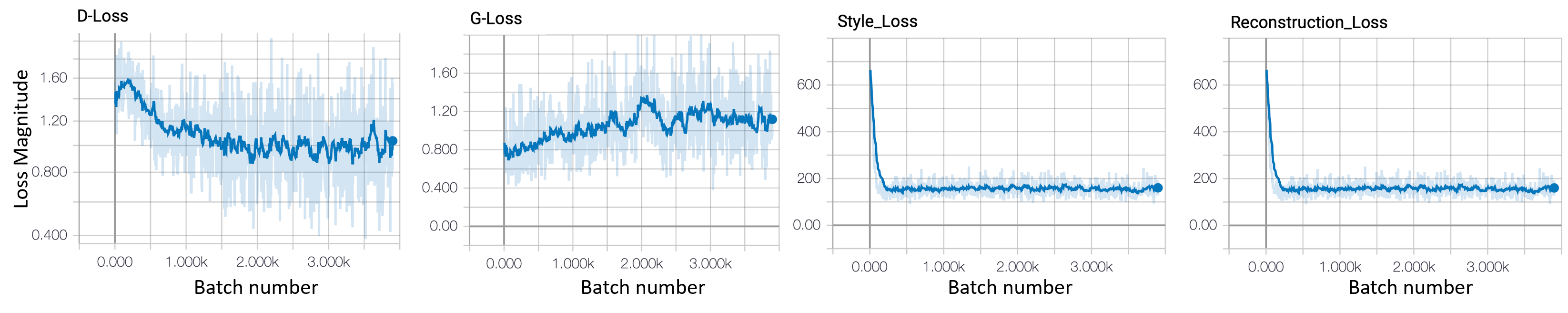}
    \caption{Loss curves over 14 epochs of the GAN without an encoder trainded with GAN, reconstruction and style loss on the CycleGAN dataset. The image quality improves after 3-4 epochs when the reconstruction and style losses start to plateau. \label{GAN_Losses}}
\end{figure}

This model performed, by far, the best, especially when trained with the reconstruction and style losses. The training curves for one of our runs is shown in figure \ref{GAN_Losses}. Some examples of maps generated using this model trained on the CycleGAN dataset can be found in figure \ref{b1}. More samples can be found in the appendix. This data is all from the test, not training dataset and was randomly sampled (not hand selected). 

\begin{figure}
	\centering
	\includegraphics[width=.6\textwidth]{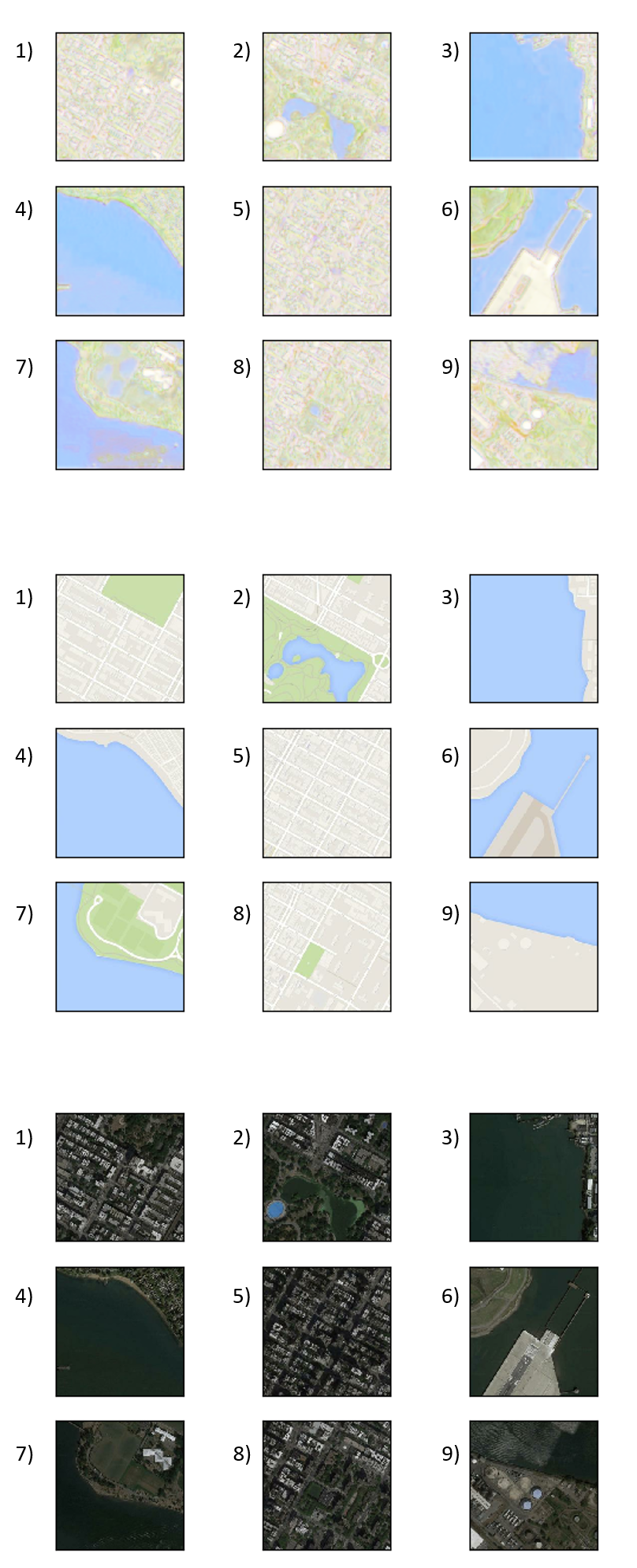}
    \caption{A sample of 9 maps generated by GeoGAN - Model 3 trained on the CycleGAN dataset. \label{b1}}
\end{figure}

\subsection{RealNVP}
We ran a sanity check on our RealNVP based model by running it through MNIST. However, it seems that the model suffers from mode collapse as the generate example all tend to match the same digit, in this case the digit 9. A figure of this is in the appendix.  We suspect that this is due to the model requiring more complexity or ill-posing the training objective. Not passing the MNIST sanity-check lead us terminate moving forward with the idea of using a RealNVP generator \ref{MNIST_Generated}. We notice the loss curves for this type of model also resemble how a GAN training on MNIST should look like.

\section{Analysis}
\subsection{Baseline Conditional GAN}
The conditional GAN with encoder was not able to generate convincing maps despite our search over hyperparameters such as size of the encoded vector, varying degrees and types of noise and slightly varying architectures. It might be that creating a meaningful embedding and generating realistic images is too demanding of a task for the Generator. We tried techniques such as having more training updates for the generator for each update of the discriminator to little success. It is possible that if we pretrained the encoder as part of an auto-encoder before plugging it in to the generator it would have better odds of success. It would be great to get the autoencoder working because then we could more naturally interpolate in this latent space to generate different maps as well as add custom features to this encoding in order to generate new maps. 

\subsection{Conditional GAN Without Encoder}
As is seen from the figures, the model trains well and converges by the end of the tenth epoch. The generator and the discriminator are only three layers deep and preliminary investigations by increasing the depth of the discriminator and generator show that the loss values are lower.  

\subsection{RealNVP}
The suspected mode collapse of the model might be due to lack of complexity since we use an FC approach to create the shift and scale function of the bisections. However, experiments with even adding more hidden layers or doubling the size of the two hidden layers quickly resulted in numerical overflow when applying the gradient updates. Additionally, the true RealNVP paper implementation used a convolutional and residual network approach for these functions for more complicated data like faces. However, there wasn't a way to add this complexity with the Tensorflow API function that applied the RealNVP transformations. Another point of conflict that might be happening is that we trained the RealNVP generator through the GAN loss functions. This is different than how flow models are are trained in practice, which is through maximum likelihood estimation on the training data. Recall that the point of the GAN generative model approach was to avoid thinking about likelihoods and instead let the adversarial examples be the training signal. Thus, it could be that the RealNVP generator learned to generate samples that might fool the discriminator at times by creating the same digit out of the noise inputs. The model however didn't learn diversity since the gradient wasn't computed with respect to the likelihood of the RealNVP generating samples from the training set. Mixing likelihood maximization and GAN based training might be an area of further research to validate that such a hybrid approach is possible.

\subsection{Data}
Our best results were achieved while training on the CycleGAN maps dataset \cite{zhu18, park_dataset} as opposed to our generated data-set because the satellite imagery was much higher resolution and more free from clouds. It would appear that satellite imagery resolutions of 10m\(^2\) per pixel does not yield great results. Our model is often confused by water vs vegetation (as both are dark blues and greens colors) and sometimes reflections on the surface of water. Therefore we think that adding IR or other bands (which are present in our generated dataset).

\section{Conclusion}
In conclusion, we demonstrate a viable model for generating maps (the GAN with reconstruction and style transfer losses and no encoder). The results shown have no hyperparameter search and are only trained on 15 epochs, we believe that with more training time we can achieve even more realistic results. Future work could also include training on more bands, such as IR to help the map distinguish between visually ambiguous regions such as lakes and vegetation. These additional bands exist in the data we pulled from google earth engine. We also provide code to generate new datasets from earth engine and google maps API. While the 10m resolution images from sentinel 2 do no train well, our data gathering streams can pull any data from google earth engine including satellite imagery, elevation data, thermal data, and more. A dataset for any translation between those domains or between maps is able to be pulled. 

\bibliographystyle{aaai}
\bibliography{MainPaper}

\appendix


\newpage
\section{MNIST Sanity Checks}
\begin{figure}[ht]
	\centering
	\includegraphics[width=\textwidth]{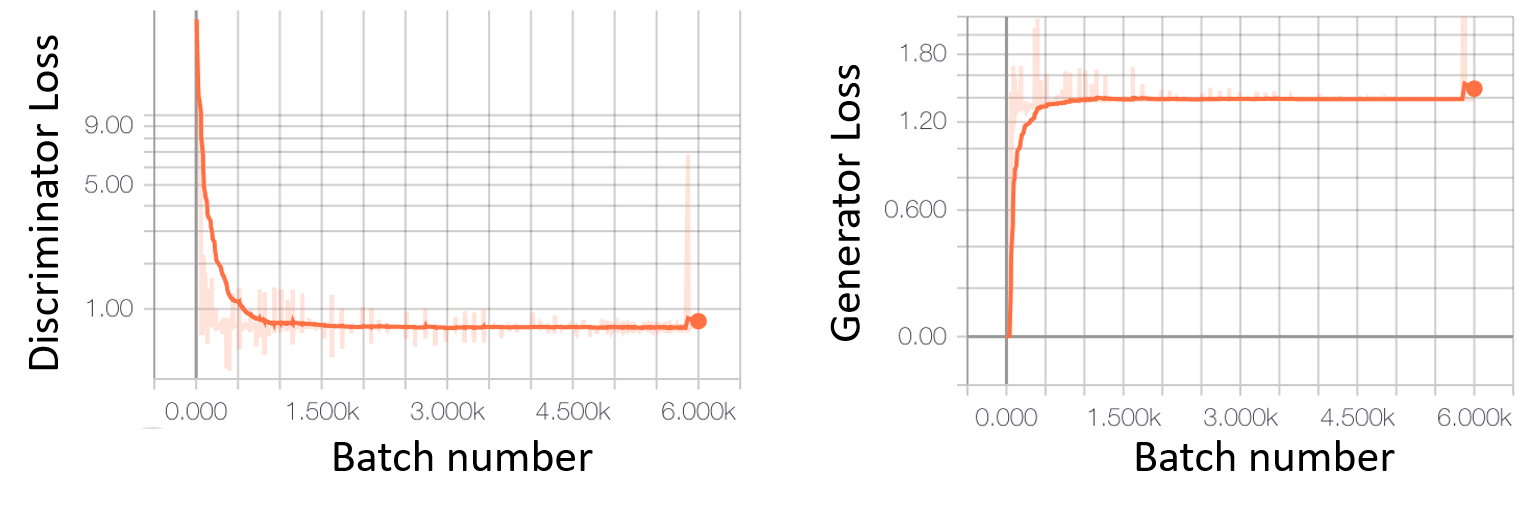}
    \caption{The discriminator and generator loss of the GeoGAN encoder trained on MNIST to evaluate whether the model can capture a simple distribution like handwritten digits.\label{mnist_loss_curves}}
\end{figure}

\begin{figure}[ht]
	\centering
	\includegraphics[width=0.45\textwidth]{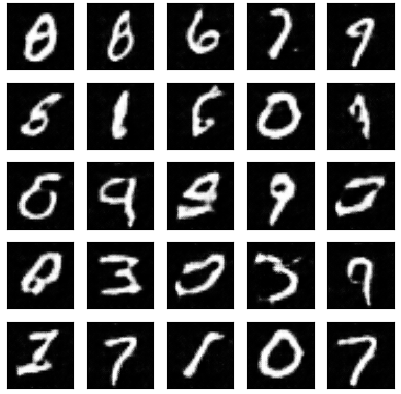}
	\includegraphics[width=0.45\textwidth]{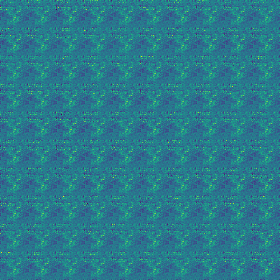}
    \caption{Generated samples from the GeoGAN at epoch 20 when trained on the MNIST dataset on the left and those from RealNVP-based generator model at 6 epochs of training on the MNIST dataset. Each small tile is a different sample. There is a clear indication of mode collapse with the RealNVP generator which we were not able to fix. \label{MNIST_Generated}}
\end{figure}

As a sanity check for our baseline architecture, we train our model on the MNIST dataset. The discriminator and generator loss over time in this case is shown in figure \ref{mnist_loss_curves}. A sample of the generated images at epoch 20 from this training and an example output of training the NVP model is shown in figure \ref{MNIST_Generated}.

\newpage
\section{Demonstration: Seasonality Effects are Encoded by Using Four Pairs (Summer, Fall, Winter, Spring) of Satellite Images for the Same Map:}
\begin{figure}[ht]
	\centering
	\includegraphics[width=\textwidth]{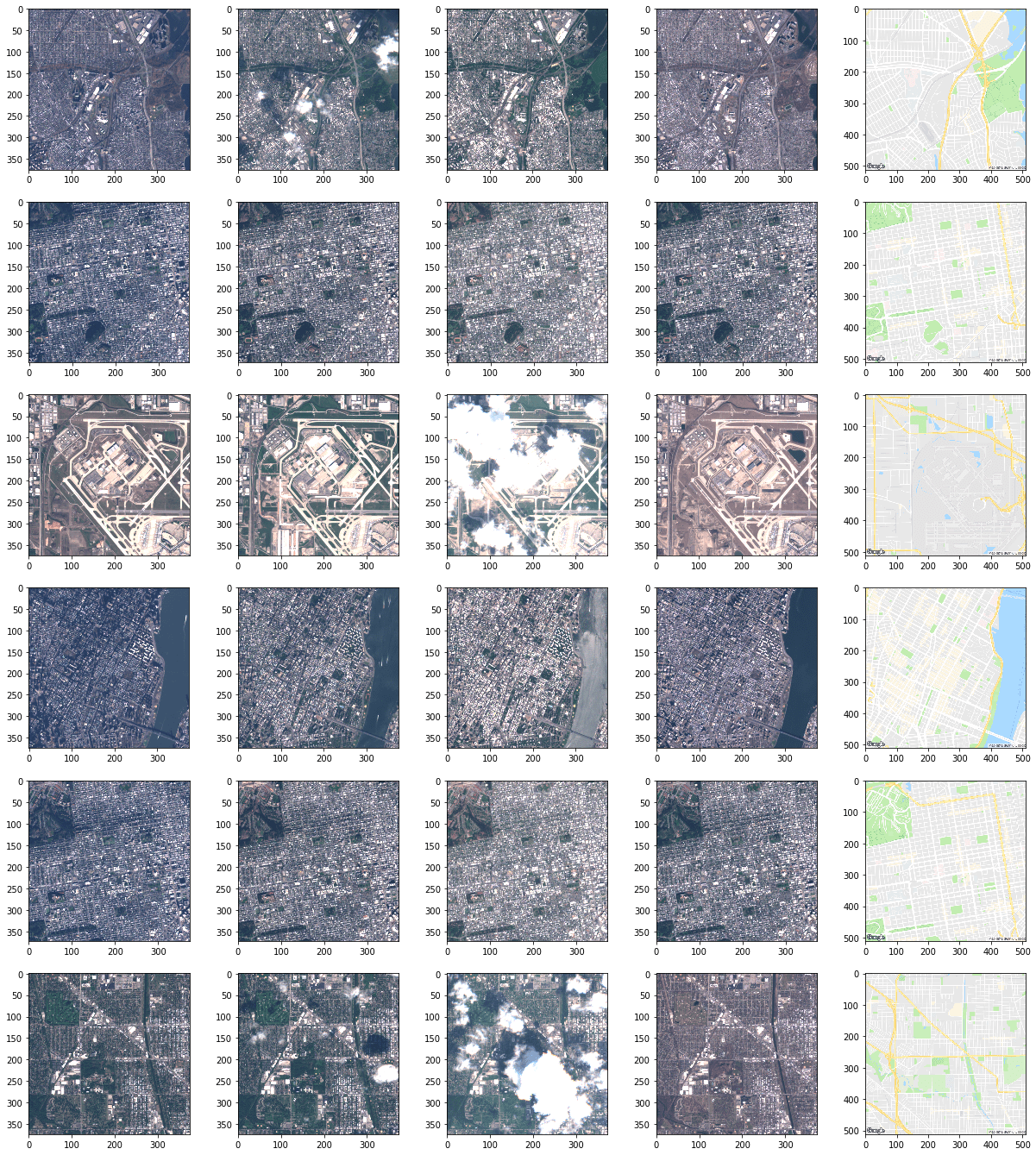}
    \caption{A grid comparing the the satellite images from Google Earth Engine of the same tile at different seasons (four seasons) along with the corresponding Google Maps tile. }
\end{figure}

\newpage
\section{Additional Samples from Trained GeoGAN (Model 3):}
\begin{figure}[ht]
	\centering
	\includegraphics[width=.5\textwidth]{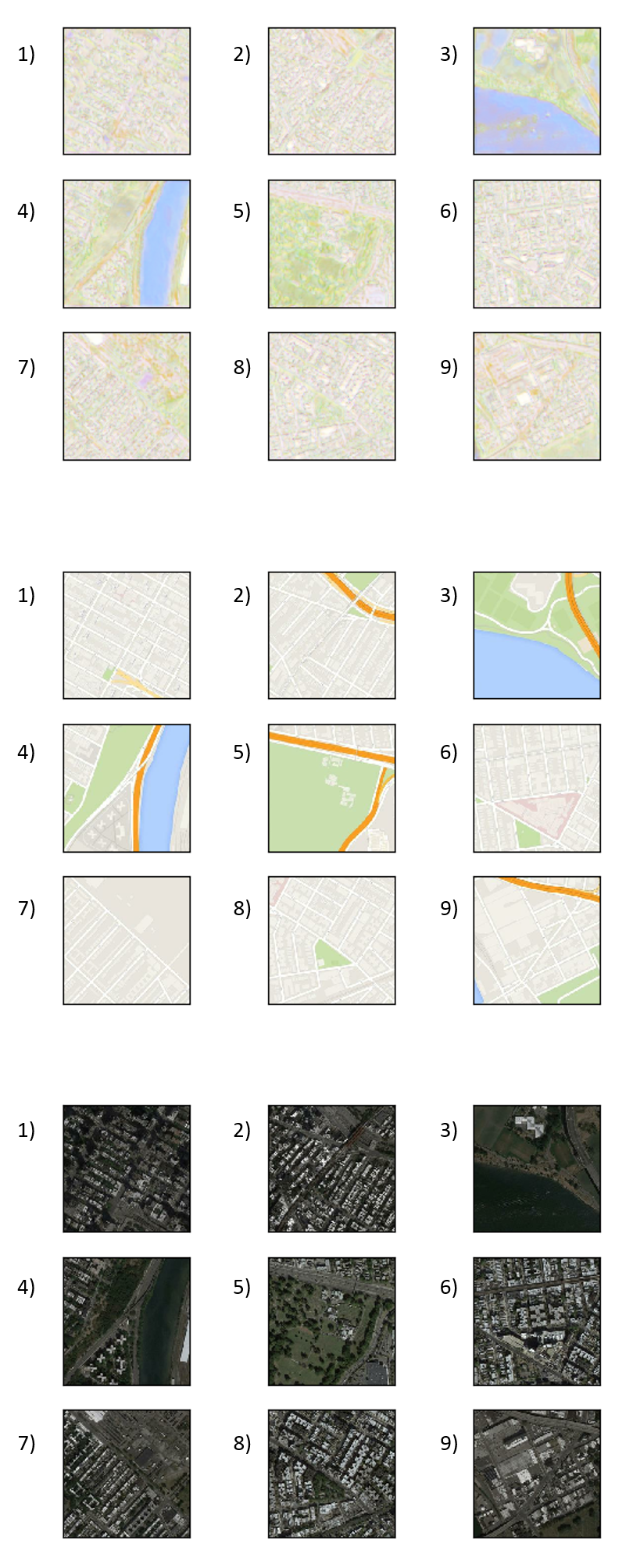}
    \caption{A sample of 9 maps generated by GeoGAN - Model 3 trained on the CycleGAN dataset.}
\end{figure}

\begin{figure}[ht]
	\centering
	\includegraphics[width=.6\textwidth]{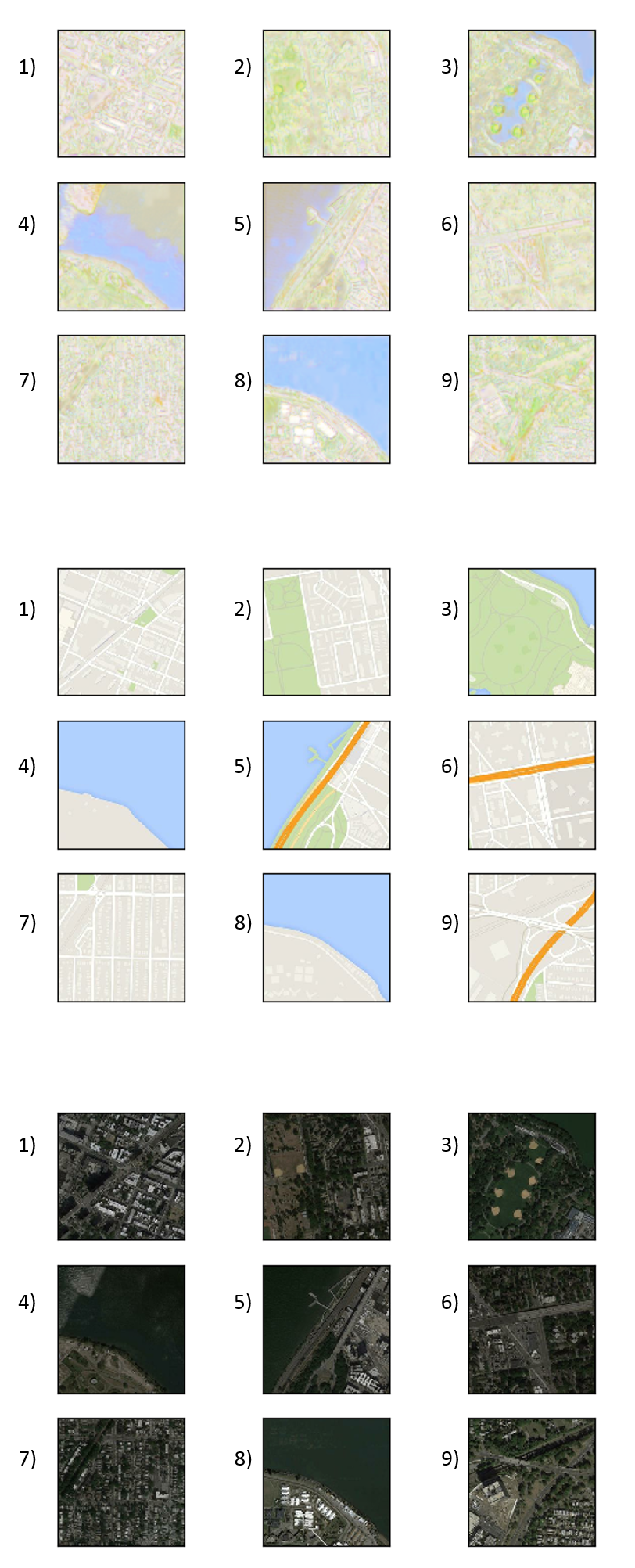}
    \caption{A sample of 9 maps generated by GeoGAN - Model 3 trained on the CycleGAN dataset.}
\end{figure}

\begin{figure}[ht]
	\centering
	\includegraphics[width=.6\textwidth]{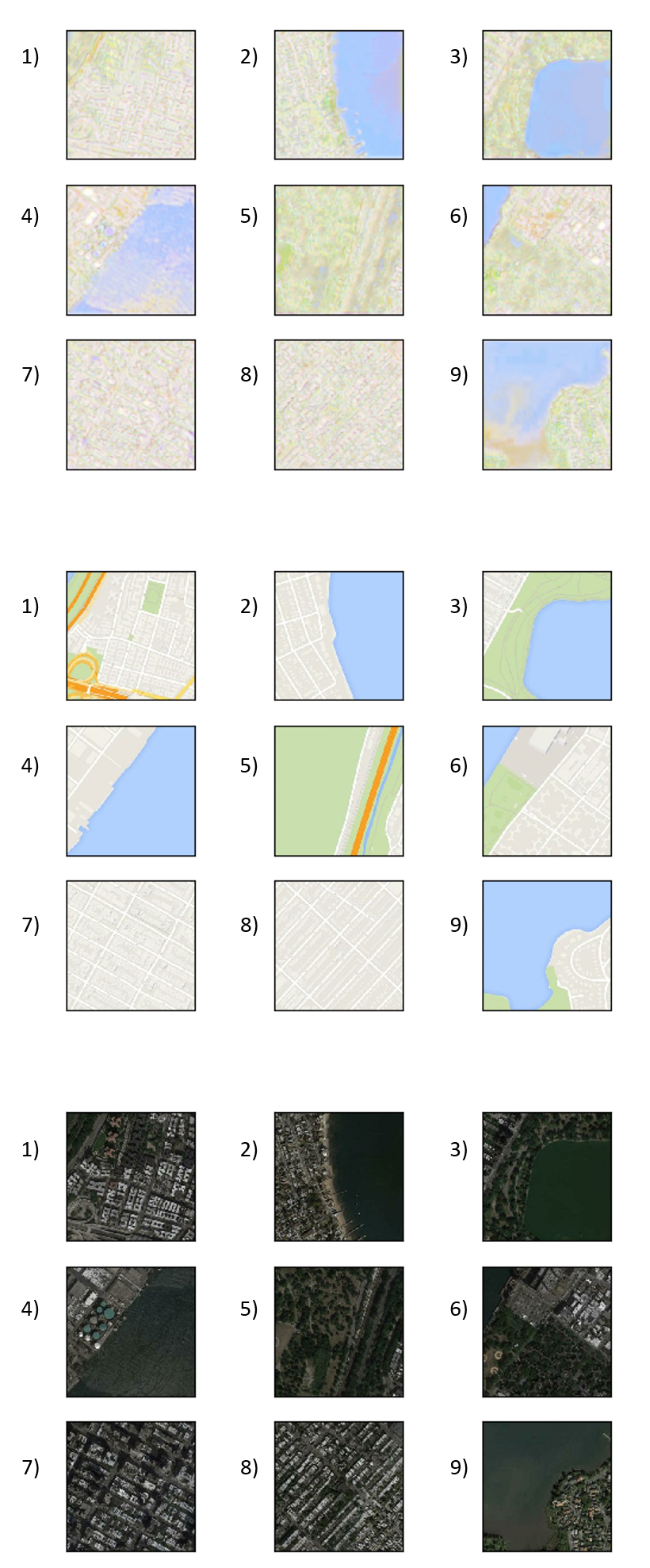}
    \caption{A sample of 9 maps generated by GeoGAN - Model 3 trained on the CycleGAN dataset.}
\end{figure}

\begin{figure}[ht]
	\centering
	\includegraphics[width=.6\textwidth]{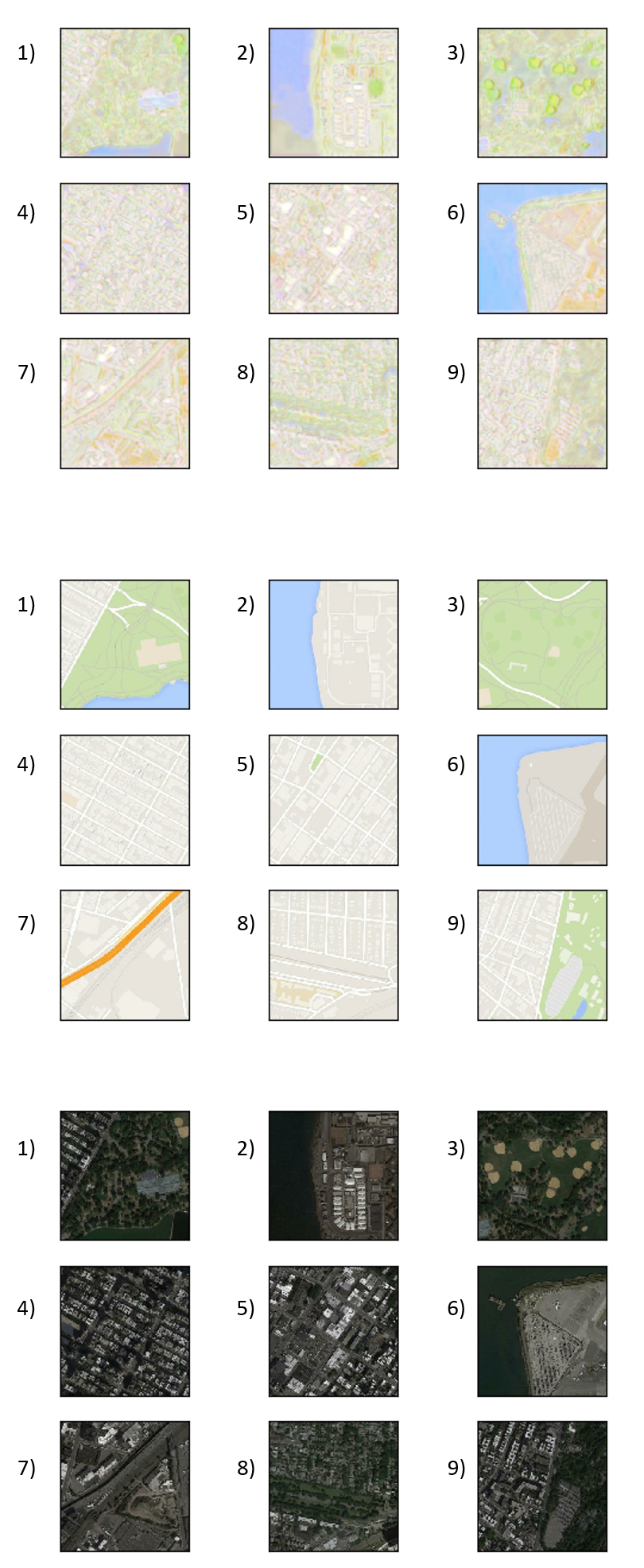}
    \caption{A sample of 9 maps generated by GeoGAN - Model 3 trained on the CycleGAN dataset.}
\end{figure}

\begin{figure}[ht]
	\centering
	\includegraphics[width=.6\textwidth]{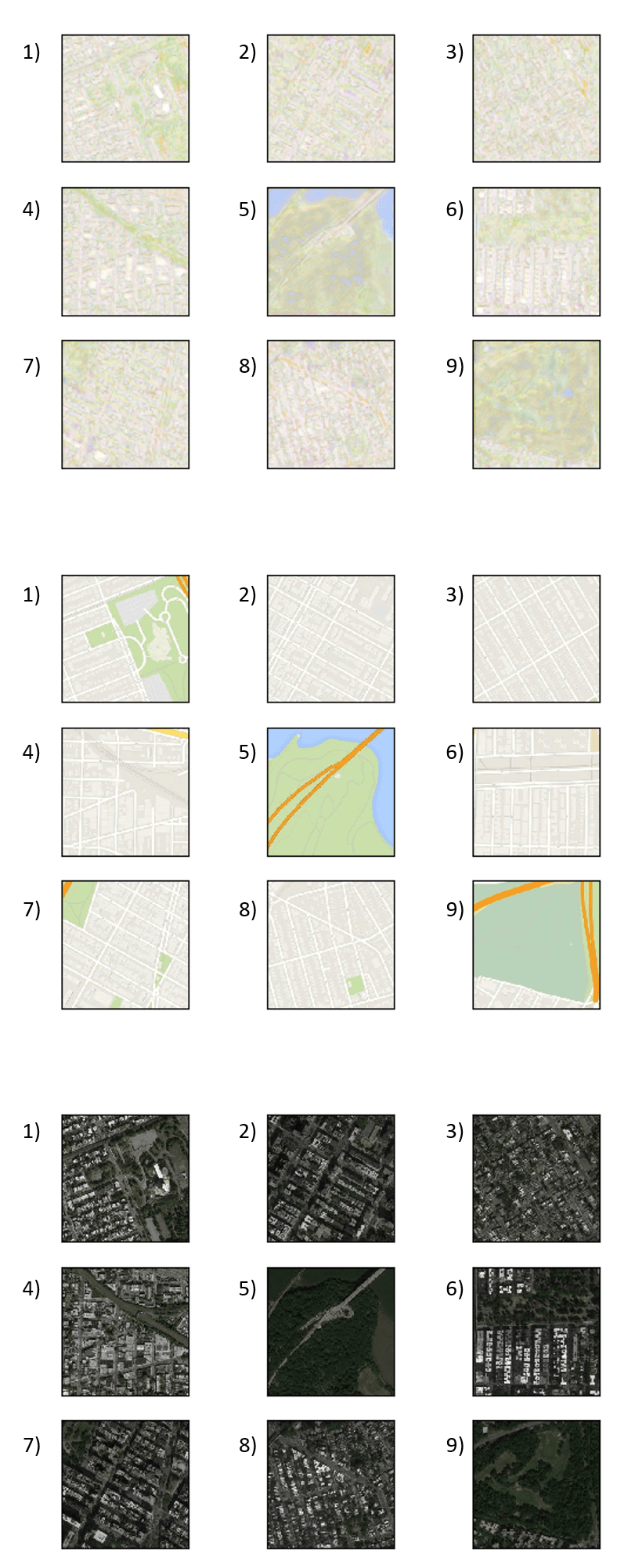}
    \caption{A sample of 9 maps generated by GeoGAN - Model 3 trained on the CycleGAN dataset.}
\end{figure}

\end{document}